\documentclass[10pt]{article}

\usepackage{placeins}
\usepackage{multicol}
\usepackage{caption}
\usepackage{amsmath}
\usepackage{amssymb}
\usepackage[utf8]{inputenc}
\usepackage[T1]{fontenc}
\usepackage{newtxtext,newtxmath}
\usepackage{times}
\usepackage[numbers,sort&compress]{natbib}
\usepackage{graphicx}
\usepackage[margin=1in]{geometry}
\usepackage{cuted}
\usepackage{setspace}
\usepackage{xcolor}
\usepackage[dvipsnames]{xcolor}
\usepackage{hyperref}
\usepackage{rotating}
\usepackage{pgfplots}
\usepackage{longtable}
\usepackage{threeparttable}
\usepackage{booktabs}
\usepackage{tabularx}
\usepackage{array}
\usepackage{multirow}
\usepackage{float}
\usepackage{stfloats}
\usepackage{authblk}
\usepackage{setspace}
\usepackage{graphicx}
\usepackage{adjustbox}
\usepackage{caption}
\usepackage{subcaption}
\usepackage{float}
\usepackage{pdflscape}
\usepackage{ragged2e}
\usepackage[table]{xcolor}
\usepackage{colortbl}
\usepackage{adjustbox}

\usepackage{titlesec}
\pgfplotsset{compat=1.18}

\titleformat{\section}{\center\normalfont\large}{\thesection}{0.3em}{}
\titlespacing*{\section}{0pt}{0.9ex}{0.9ex}

\titleformat{\subsection}{\normalfont\normalsize\itshape}{\thesubsection}{0.3em}{}

\titlespacing*{\subsection}{0pt}{1ex}{1ex}

\titleformat{\subsubsection}{\normalfont\small\itshape}{\thesubsubsection}{0.3em}{}
\titlespacing*{\subsubsection}{0pt}{1ex}{1ex}

\renewcommand{\thesection}{\arabic{section}.}
\renewcommand{\thesubsection}{\thesection\arabic{subsection}}
\renewcommand{\thesubsubsection}{\thesubsection.\arabic{subsubsection}}

\makeatletter
\renewcommand{\p@subsection}{}
\renewcommand{\p@subsubsection}{}
\makeatother
\hypersetup{
    colorlinks=true,
    citecolor=blue,
    linkcolor=blue,
    urlcolor=blue
}
\begin{document}
{\bfseries\Large
\begin{center}
Evaluating Reliability in Machine Learning Models for Early\\
Chronic Kidney Disease Prediction: A Systematic Review of\\
Data Leakage and Predictor Stability
\end{center}
}

\vspace{0.5em}
\begin{center}
Mashrul Hossain, Nafesa Kibria, Fahim Shahriar \\
\vspace{0.3em}
\texttt{mashrul16hossain@gmail.com, nafesa220128@gmail.com, fahimshahriar1306@gmail.com}
\\
\vspace{0.3em}
East West University, Dhaka, Bangladesh
\end{center}

\noindent
\noindent
\vspace{0.1em}
\noindent
{\fontsize{10}{13}\selectfont\textbf{
ABSTRACT - }
{\fontsize{10}{13}\selectfont{The early detection of Chronic Kidney Disease using machine learning has attracted significant interest in healthcare-related computer science. Despite rapid advancements in this field, many reported studies remain inconsistent and potentially misleading. A significant drawback is the lack of organized evaluation regarding methodological concerns. Key issues include data leakage, limited access to temporal patient records and inconsistency in reported clinical indicators. This research offers a systematic literature review of existing CKD prediction studies using interpretable machine learning techniques, where nineteen relevant studies were selected via systematic searches across major academic databases. To assess methodological reliability, this study introduces a structured taxonomy of information leakage and a quantitative leakage scoring framework to systematically evaluate reliability across CKD prediction studies.
The analysis reveals a strong relationship between leakage and inflated performance. Here, High leakage-studies report an average accuracy of 95.48\%, compared to 80.2\% for leakage-free studies, reflecting an increase of approximately 15.28\%. Furthermore, a cross-study feature stability analysis shows that only a small subset of predictors is consistently reproducible, with over 80\% lacking reliability. Overall, the findings suggest that many reported performance improvements stem from methodological limitations rather than true predictive capability.
}
}
\vspace{0.4em}
\\
\noindent
{\fontsize{9}{10}\selectfont
\section{Introduction}
\noindent
{\fontsize{10}{12}\selectfont
Chronic Kidney Disease (CKD) is a progressive condition characterized by gradual decline in kidney function, affecting more than 850 million individuals worldwide \cite{1.5}\cite{850m}. Early detection of CKD is crucial because symptoms often appear only in advanced stages when treatment options  quite limited. Early identification of CKD is therefore essential to delay disease progression, manage associated comorbidities and reduce the long-term economic burden on healthcare systems \cite{1.6}\cite{2404}. In recent years, machine learning (ML) has emerged as a promising approach for analyzing complex clinical datasets and identifying patterns associated with the early onset of CKD that may not be detected through conventional statistical techniques \cite{1.1}\cite{12}.

Machine Learning models can evaluate various types of patient data, such as lab results, demographic information and medical history, to assess a person's likelihood of developing CKD. These forecasting systems assist healthcare providers in identifying high-risk patients sooner and facilitating preventive treatment strategies before significant kidney damage occurs \cite{1.6}\cite{12}. However, numerous critical challenges continue to restrict the real world implementation of ML driven diagnostic systems in healthcare. A well known problem is the restricted interpretability associated with numerous high-performing models, such as deep neural networks and complex ensemble techniques \cite{1.6}\cite{2.6}. Clinicians frequently require clear justifications for algorithmic predictions to trust them and integrate them into their medical decision making. Also, methodological limitations such as dataset bias, limited external validation and small sample sizes from individual centers continue to affect the reliability and generalizability of many existing studies\cite{1.1}\cite{12}\cite{5.5}.

Recent studies on early CKD prediction indicate a gradual shift from traditional statistical approaches to more advanced machine learning methods. Logistic regression is frequently utilized as a straightforward baseline model, while ensemble methods based on trees such as Random Forest, XGBoost and CatBoost which are typically applied to improve prediction accuracy \cite{1.1}\cite{2.3}\cite{19.6}. These models usually depend on several widely cited clinical indicators, such as serum creatinine concentrations, estimated glomerular filtration rate (eGFR), hemoglobin concentrations, markers of albuminuria and demographic variables including age or hypertension history \cite{1.1}\cite{1.5}\cite{4.4}\cite{19.6}. Multiple studies demonstrate exceptionally high diagnostic precision and robust area-under the curve (AUC) metrics while evaluating their models on selected datasets. Consequently, models that achieve almost flawless outcomes in controlled studies may not function as effectively in real healthcare settings, where data quality, patient diversity and clinical variability are considerably higher\cite{12}\cite{AsystematicReview}.

Despite the rapid advancements in machine learning research for early CKD prediction, several methodological issues remain insufficiently addressed in the literature. A key issue in CKD detection studies is the information leakage, which occurs when predictors contain information directly tied to the outcome label used during model training. For example, variables such as eGFR or serum creatinine are sometimes used as predictors even when CKD labels are derived from the same measurements, which can lead to overly optimistic model performance \cite{1.4}. Again, the predictors which is identified as important often differ across studies, suggesting that variability in feature significance is based on feature selection technique and dataset traits \cite{1.4}. Although previous studies on CKD prediction mainly compared model performance, they lack a unified framework to systematically detect information leakage or evaluate cross-study feature stability. This limitation highlights the need for a comprehensive evaluation of methodological validity, consistency of predictors and clinical relevance in existing CKD studies.
\vspace{0.4em}
\noindent
\\
\textbf{Research Questions :}
\\
• How much does the data leakage influence ML models effectiveness?
\vspace{0.2em}
\\
\noindent
• Which clinical features show consistent stability in early CKD prediction studies and which predictors are dataset specific?
\vspace{0.2em}
\\
\noindent
\textbf{Objectives :}
This study addresses the mentioned challenges by providing a structured analysis of methodological approaches in machine learning based CKD prediction research. The primary objectives of this study are outlined as follows:
\noindent
\\
  • A categorization of information leakage in CKD machine learning study, classifying direct leakage, proxy leakage and temporal leakage to provide more consistent experimental design in future studies.
\\
  • Cross-study feature stability analysis in three steps to identify clinical predictors that remain consistent across multiple populations and those that seems to be dataset-specific predictors.
\section{Related Works}
{\fontsize{10}{12}\selectfont
 Detecting of Chronic Kidney Disease early is vital as it can slow down the disease progression. CKD doesn't
 show symptoms in initial stages, so most cases are found in primary care settings\cite{1.6,AsystematicReview}. Machine learning interpretable frameworks has emerged as a significant tool to sort complex clinical unorganized data and uncover patterns for risk detection that experienced clinicians may miss \cite{12,AsystematicReview}. Early CKD prediction studies relied on classical algorithms, such as Logistic Regression, Support Vector Machines and Decision Trees \cite{2.earlypred,lr-1}. These models  generalizability was severely limited because of small, single-center datasets and minimal data quality management \cite{13.6,AsystematicReview}. Specifically, the high performance scores often reported on the UCI benchmark serve as evidence of overfitting, raising doubts about model reliability in real-world healthcare settings. In real-world with real diversity in patients and clinical chaos, these models would likely stumble. \cite{12,AsystematicReview}. There is methodological agreement regarding the importance of essential biomarkers such as serum creatinine and estimated glomerular filtration rate (eGFR), often identified through feature selection methods like Recursive Feature Elimination (RFE), Principal Component Analysis (PCA) or LASSO \cite{3.5,AsystematicReview}. Furthermore, common predictors are albuminuria, patient age and hemoglobin levels \cite{1.4,nefroai}. A core methodological concern is the instability of leading predictors across studies. These variations often reflect dataset biases and regional demographic differences such as varying environmental factors and health system constraints \cite{1.4,d6}.

A critical limitation identified is the widespread presence of information leakage. Numerous studies include pathology-test-based markers, such as serum creatinine or eGFR, as input features for model training even though those same exact tests are used to define CKD in the first place \cite{1.4}. This practice introduces circular reasoning, where model ends up just matching the abnormal values instead of actually predicting anything useful, which results in misleading performance metrics that render the model redundant for clinical screening \cite{1.4,lr-1}. To address this issue, recent work focuses on leakage-safe preprocessing pipelines, ensuring that operations like scaling, outlier capping and class balancing are performed strictly within the training split \cite{1.5}. Also, Performance evaluation has shifted towards metrics that handle imbalanced data better, like AUC-ROC, F1-score and Recall \cite{1.1,1.4}. 
\section{Methodology}
\subsection{Literature Search Strategy}
{\fontsize{10}{12}\selectfont
A detailed search was conducted to identify studies on the early detection of Chronic Kidney Disease utilizing machine learning methods. Searching through many academic databases, such as IEEE Xplore, PubMed and Google Scholar, several papers were selected in the first stage. 

The search utilized different combinations of particular keywords. Essential terms included: “CKD" or Chronic Kidney Disease” or “CKD early detection” or “machine learning” or “ML”, “clinical indicators” or “clinical attributes”. These were combined into queries such as (“CKD” AND “machine learning” AND “early detection”) and (“chronic kidney disease” AND “clinical predictors”).

The search focused on research published from 2021 to 2025 to include recent advancements in machine learning. Additionally, backward and forward citation tracking was conducted to discover relevant studies beyond the original search results.
\subsection{Study Selection Process}
The selection of studies was conducted using a systematic method based on the PRISMA framework. Studies were first collected from various databases employing the specified search strategy. Duplicate entries were eliminated and the remaining documents were subjected to title and abstract screening to evaluate their relevance.
Potentially relevant studies were then subjected to full-text review. During this stage, papers were assessed based on their emphasis on early-stage CKD detection, application of machine learning methods, inclusion of clinical predictors and incorporation of interpretability or explainability techniques.
A total of 494 records were identified. After screening, reducing irrelevant studies, 19 studies were selected.
\noindent
Figure \ref{fig:prisma} is the visual representation of the study selection process. This structural filtering ensures that only relevant and high-quality studies are considered.
\begin{figure}[!t]
    \centering
    \includegraphics[
    width=0.72\linewidth,
    keepaspectratio
    ]{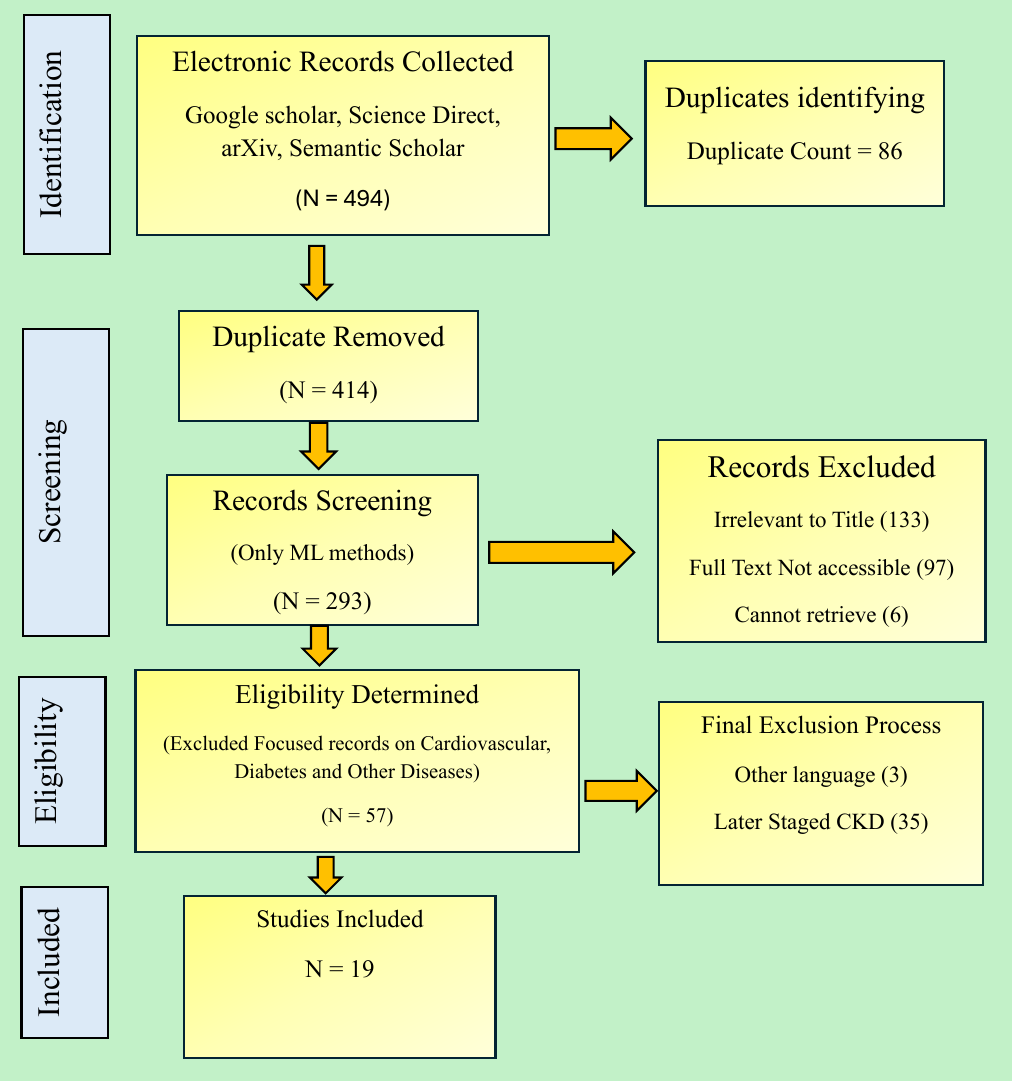}
    \caption{PRISMA flowchart for study selection process}
    \label{fig:prisma}
\end{figure}
\subsection{Inclusion and Exclusion Criteria}
Explicit inclusion and exclusion criteria were defined to maintain consistency during study selection.
\subsubsection{Inclusion Criteria}
• Focused on ML-based early CKD prediction\\
• Addressed early stage prediction\\
• Utilized clinical features (laboratory or demographic data)\\
• Published in English within (2021-2025)
\noindent
\subsubsection{Exclusion Criteria}
• Focused only on medical imaging without predictors\\
• Addressed other diseases (cardiovascular, diabetes, etc)\\
• Focused on late-stage or general CKD detection\\
• Lacked clinical predictor analysis
\subsection{Datasets Overview}
Datasets used in early chronic kidney disease (CKD) research range from relatively small hospital-based clinical records to very large national repositories such as the UK Biobank, which contains health information from more than half
\begin{table*}[!t]
\footnotesize
\centering
\caption{Summary of Datasets Used Across Reviewed Studies}
\label{tab:datasets}

\begin{tabular}{p{3cm}p{1.7cm}p{1.2cm}p{2.15cm}p{1.51cm}p{1.4cm}p{3.2cm}}
\toprule
\textbf{Dataset}&\textbf{Country} &\textbf{Sample Size}&\textbf{CKD Cases (\%)}&\textbf{CKD Stages}&\textbf{Features} &\textbf{Ref} \\
\midrule

UCI CKD Repo & India & 400 & 250 (62.5\%) & - & 24-26 & \cite{1.5,13.6,6.6,11.4,1.6,8.1,3.5,6.3} \\

NHANES & USA & 28,377 & Varies & Stages 1--5 & 36 & \cite{6.5} \\

UK Biobank & UK & 15,811 & CKD cohort & Stages 1--5 & 22 & \cite{5.4} \\

BD Community & Bangladesh & 284 & 112 (39.4\%) & Stages 1--3 & 24-29 & \cite{1.4} \\

TH Dataset & UAE & 491 & 56 (11.4\%) & Stages 3--5 & 21 & \cite{1.4,2.6} \\

Korean Multi-Center & South Korea & 5,120 & 1,361 (26.6\%) & - & 15 & \cite{15} \\

Thai T2DM Hospital & Thailand & 3,471 & 53.7\% & eGFR < 60 & 8 & \cite{5.6} \\

Chinese T2DM & China & 12,190 & 1,391 (14.3\%) & - & 32 & \cite{5} \\
Synthetic CKD Cohort & N/A & - & 55\% & -
& (24) & \cite{1.5} \\

Bab Al-Rayan Clinic & Sudan & 100 & - & - & Varies (8) & \cite{12} \\

Enam Medical College & Bangladesh & 200 & - & Stages 1–5 & 28 & \cite{3.5} \\

China Multicenter Study & China & 22,263 & 11,667 & (Stages 1-2)
& - & \cite{identification}
\\
\bottomrule
\end{tabular}
\end{table*}
 a million participants \cite{5.4}. Among the datasets commonly used for benchmarking machine learning models, the UCI Machine Learning Repository remains one of the most widely adopted sources for algorithm evaluation and comparison \cite{1.4,11.4,6.6}. At the same time, recent studies increasingly emphasize community-based screening datasets, particularly in South Asian regions, to support early detection strategies in low-resource settings where access to advanced laboratory testing is limited \cite{1.4,7.4}.

Real-world clinical datasets contain high-dimensional feature spaces, irregular time intervals between medical observations and substantial class imbalance, where CKD-positive cases represent a minority of the population \cite{1.1,11.4,6.3}. Longitudinal Electronic Medical Record (EMR) data further highlight the dynamic nature of disease progression. For instance, evidence suggests that approximately 20.3\% of diabetic patients showing early kidney impairment develop CKD within six months\cite{1.1}. Most predictive modeling studies rely on datasets containing roughly 24 to 36 clinical attributes, which include demographic factors as well as biochemical markers such as serum creatinine and estimated glomerular filtration rate (eGFR) \cite{11.4,4.4,6.5}.

However, widely used benchmark datasets may not fully represent the demographic diversity of global populations. As a result, several studies validate their models using independent cohorts collected from countries such as Bangladesh, Korea, China and the United Arab Emirates \cite{1.4,15,5,2.6}. National health surveys like NHANES are also frequently used because they follow standardized data collection procedures and incorporate additional lifestyle variables, including dietary patterns and sleep habits, which can influence kidney health outcomes\cite{4.4,6.5}. Table~\ref{tab:datasets} summarizes the
datasets used across reviewed studies.

\subsection{Preprocessing Techniques}
Data preprocessing is a crucial step in preparing healthcare datasets for machine learning analysis, as raw clinical data often contain inconsistencies and missing values\cite{2.6,6.5}. One of the most common challenges involves handling incomplete records, which may arise from random data loss or manual entry errors during clinical documentation \cite{6.6,11.4}. To address this issue, researchers frequently apply imputation techniques such as Multivariate Imputation by Chained Equations (MICE), K-nearest neighbor (KNN) imputation or median replacement to reduce the influence of missing information while maintaining robustness against outliers\cite{1.4,11.4,4.4}.

Additionally to handle temporal variability, longitudinal medical records are frequently summarized within a specific observation window using statistical measures such as the mean, quartiles or standard deviation\cite{1.1}. Feature scaling is also routinely performed to ensure consistent data ranges across variables, typically through Min–Max normalization or Z-score standardization\cite{1.4,11.4,7.4,6.6}. Moreover, techniques for identifying and correcting outliers like interquartile range (IQR) are employed to enhance model stability and prediction accuracy\cite{7.4,11.4}. Due to the common issue of class imbalance in CKD datasets, methods such as the Synthetic Minority Over-sampling Technique (SMOTE) or random oversampling are frequently employed to equilibrate the distribution between healthy and CKD-affected cases\cite{11.4,6.3,5.4}. 
\subsection{Feature Selection Methods}
Feature selection is equally crucial in determining the most informative clinical predictors. These techniques are typically categorized into statistical filter methods, model-based wrapper approaches and regularization strategies\cite{1.4,7.4}. Tests like Pearson correlation, ANOVA and the Mann–Whitney U test are frequently utilized to identify variables that significantly vary between CKD and non-CKD populations\cite{1.4,11.4,5.4}. Wrapper-based methods like Recursive Feature Elimination with Cross-Validation (RFECV) and Sequential Feature Selection (SFS) have demonstrated the ability to discover smaller subsets of features that often outperform models trained on the complete set of variables \cite{1.4,7.4}.
\begin{table}[!b]
\centering
\footnotesize
\setlength{\tabcolsep}{4pt}

\begin{threeparttable}
\caption{Leakage Taxonomy Definition}
\label{tab:leakage_taxonomy}

\begin{tabularx}{\textwidth}{p{3cm} p{3cm} X X}
\toprule
\textbf{Leakage Type} & \textbf{Formal Definition} & \textbf{Example Features} & \textbf{Detection Rule} \\
\midrule

Type 1 — Direct Leakage (DL) & 
Feature $\subseteq$ Diagnostic Criterion & 
eGFR, Serum Creatinine (SCr), Albuminuria & 
Check if feature defines the ground truth label \\

Type 2 — Proxy Leakage (PL) & 
$|r| \geq 0.85$ or clinical redundancy with diagnostic biomarkers & 
Blood Urea Nitrogen (BUN), Cystatin C & 
Evaluate correlation with diagnostic biomarkers \\

Type 3 — Temporal Leakage (TL) & 
Timestamp(feature) $>$ Timestamp(outcome) & 
Post-diagnosis medications, dialysis codes & 
Verify feature occurs after diagnosis timestamp \\
\bottomrule
\end{tabularx}

\end{threeparttable}
\end{table}
\noindent
\section{Data Leakage Evaluation}
The following section provides a structured evaluation of data leakage in early chronic kidney disease (CKD) prediction studies. To ensure consistent evaluation across studies, three types of data leakage are defined using the following criteria  :
\\
Type 1—A feature is considered Direct Leakage (DL) if it is included in diagnostic criteria for defining CKD. Here, the criteria consist of estimated glomerular filtration rate, serum creatinine and albuminuria. Including these as predictive features may lead to circular reasoning or may even hamper the model's performance, because these features directly impact the clinical understanding of CKD.
\noindent
\vspace{0.3em}
\\
Type 2—Proxy Leakage (PL) occurs when a feature is not part of the diagnostic criteria but is statistically or clinically redundant with one.
In this study, proxy leakage is identified using a hybrid criterion:
(i) strong statistical correlation with diagnostic biomarkers ($|r| \geq 0.85$) or
(ii) clinical or physiological dependency on the same underlying renal function pathway, even when pairwise correlations are moderate due to dataset-specific factors. The ($|r| \geq 0.85$) threshold is commonly adopted as a conservative indicator of strong association in clinical datasets.
\noindent
\vspace{0.3em}
\\
Type 3—Temporal leakage occurs when predictors gathered after the outcome event are utilized in training the model. Examples include medications after diagnosis, dialysis treatments or subsequent interventions. Since these variables are unavailable at prediction time, their inclusion leads to unrealistic model performance, a phenomenon widely recognized in machine-learning evaluation pipelines \cite{kapoor2023leakage}.
\\
Table \ref{tab:leakage_taxonomy} presents an overview of the definitions and detection criteria for every leakage category utilized in this analysis.

To quantify the extent of leakage in each study, a unified leakage score is defined as:
\begin{equation}
L_i = w_1 DL_i + w_2 PL_i + w_3 TL_i
\end{equation}

where $L_i$ represents the total leakage score for study $i$, while $DL_i$, $PL_i$ and $TL_i$ denote direct, proxy and temporal leakage components, respectively.
\vspace{0.4em}
\\

The individual leakage components are defined as:
{\small
\begin{equation}
DL_i =
\begin{cases}
1, & \text{if diagnostic features are included} \\
0, & \text{otherwise}
\end{cases}
\end{equation}
}
{\small
\begin{equation}
PL_i =
\begin{cases}
0, & |r| < 0.6 \\
1, & 0.6 \leq |r| < 0.85 \\
2, & |r| \geq 0.85
\end{cases}
\end{equation}
}
A correlation of $|r| \geq 0.85$ corresponds to a Variance Inflation Factor exceeding $6.7$ ($VIF = 1/(1-r^2)$), indicating near-collinear redundancy with diagnostic biomarkers and warranting full proxy leakage assignment ($PL = 2$) \cite{hair2019multivariate, obrien2007caution}. The intermediate range ($0.60 \leq |r| < 0.85$, $VIF = 1.6$--$6.7$) captures moderate statistical dependence that may not individually constitute leakage but creates meaningful predictive overlap, particularly when multiple such variables are included simultaneously \cite{steyerberg2010assessing} ($PL = 1$), consistent with evidence that partial information leakage can still inflate model performance \cite{kapoor2023leakage}. Variables with $|r| < 0.60$ are treated as statistically independent of diagnostic criteria ($PL = 0$), following conventional interpretations of correlation strength where values near $r = 0.50$ represent large effects in behavioral and clinical data \cite{cohen2013statistical}.
{\small
\begin{equation}
TL_i =
\begin{cases}
1, & \text{when feature timestamp crosses outcome timestamp} \\
0, & \text{All other case}
\end{cases}
\end{equation}
}

Based on the relative severity of each leakage type, weights are assigned as:

\begin{equation}
w_1 = 2, \quad w_2 = 1, \quad w_3 = 2
\end{equation}

Thus, the final leakage scoring function is:

\begin{equation}
L_i = 2DL_i + PL_i + 2TL_i
\end{equation}
\noindent
{\small
\textbf{Leakage Classification Rule}
}
\noindent
\vspace{0.7em}
\\
To facilitate interpretation, studies are categorized according to their leakage score:
{\small
\begin{equation}
\text{Leakage Category} =
\begin{cases}
\text{Low Leakage}, & L_i \leq 2 \\
\text{High Leakage}, & L_i \geq 3
\end{cases}
\end{equation}
}

\begin{figure}[htbp]
        \centering
        \includegraphics[
        width=0.91\linewidth,
        keepaspectratio
        ]{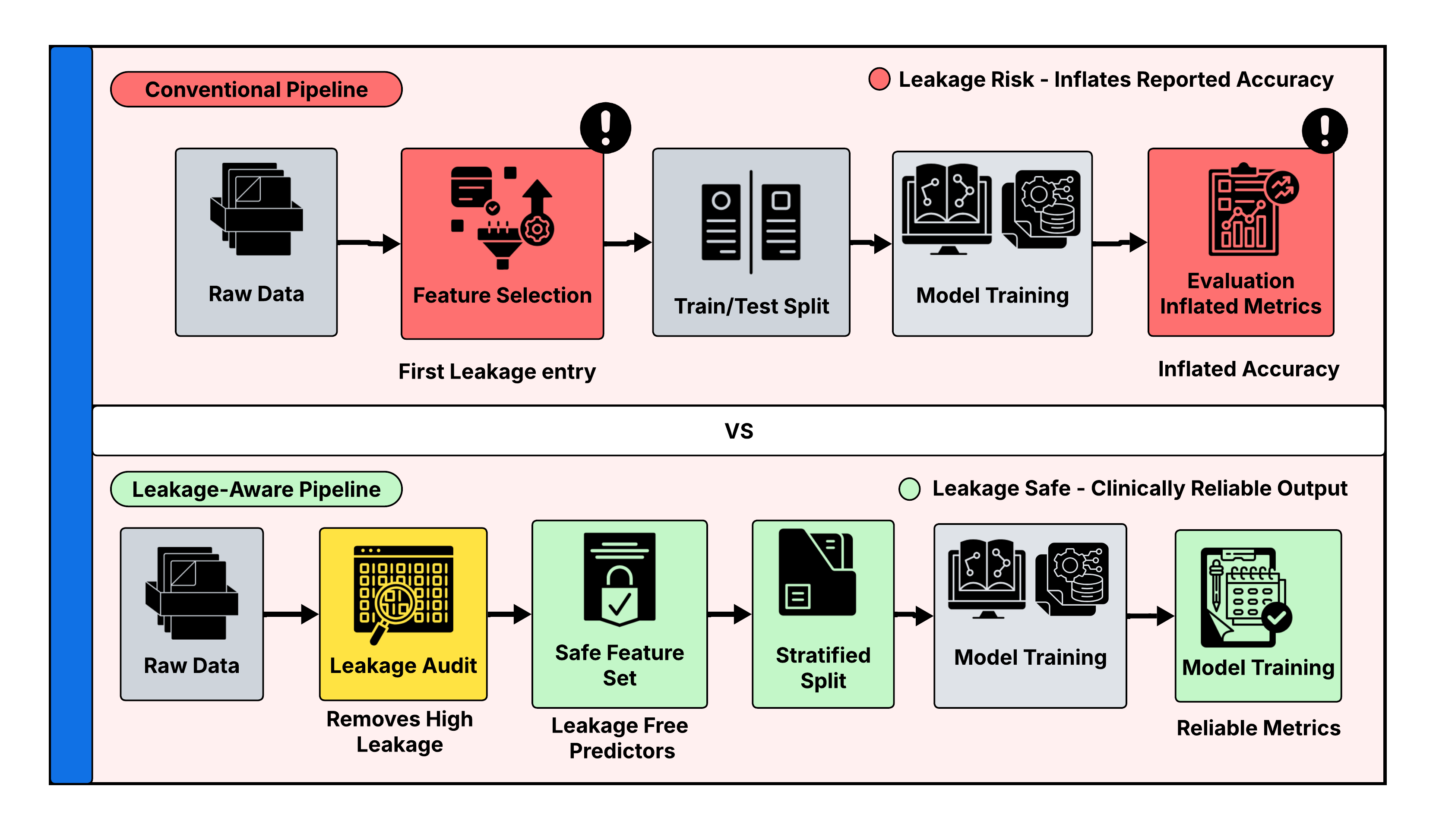}
        \caption{Conventional vs Leakage aware Pipeline}
        \label{fig:conv_vs_leak}
\end{figure}
\noindent
Figure \ref{fig:conv_vs_leak} presents a conceptual comparison between conventional and leakage-aware CKD prediction workflows. The figure highlights the importance of appropriate feature handling and leakage control to ensure reliable model evaluation.

\begin{table}[t]
\centering
\small
\caption{Performance Comparison under Controlled Leakage Injection}
\label{tab:leakage_experiment}

\begin{tabular}{p{2.5cm} p{2cm} c c c}
\hline
\textbf{Model} & \textbf{Setup} & \textbf{Acc (\%)} & \textbf{F1 (\%)} & \textbf{AUC (\%)} \\
\hline

\multirow{3}{*}{CatBoost}
& Clean  & 96.25 & 96.90 & 98.60 \\
& Proxy  & 97.50 & 97.90 & 100.00 \\
& Direct & 100.00 & 100.00 & 100.00 \\

\hline

\multirow{3}{*}{Stacked Ensemble}
& Clean  & 96.25 & 97.00 & 98.70 \\
& Proxy  & 97.50 & 98.00 & 100.00 \\
& Direct & 100.00 & 100.00 & 100.00 \\

\hline

\multirow{3}{*}{LightGBM}
& Clean  & 97.50 & 97.90 & 99.40 \\
& Proxy  & 97.50 & 97.90 & 99.93 \\
& Direct & 98.75 & 98.90 & 100.00 \\

\hline

\multirow{3}{*}{AdaBoost}
& Clean  & 92.50 & 93.61 & 98.50 \\
& Proxy  & 97.50 & 97.90 & 100.00 \\
& Direct & 100.00 & 100.00 & 100.00 \\

\hline
\end{tabular}
\end{table}
\begin{table}[!b]
\centering
\small
\setlength{\tabcolsep}{5pt}
\renewcommand{\arraystretch}{0.97}
\caption{Data Leakage Assessment Across Reviewed Studies}
\label{tab:leakage_assessment}

\begin{tabularx}{\columnwidth}{p{0.6cm}p{1.2cm}p{1.2cm}p{1.2cm}p{1.4cm}p{8.8cm}}
\toprule
\textbf{Ref}& 
\textbf{DL (0/1)}& 
\textbf{PL (0-2)}& 
\textbf{TL(0/1)}& 
\textbf{Leakage Score}& 
\textbf{Justification}\\
\midrule

\cite{1.4} & 0 & 0 & 0 & 0 & Features include hypertension, age, anemia, diabetes. 
No eGFR, SCr or ACR. Cross-sectional design. \\

\cite{11.4} & 1 & 2 & 0 & 4 &
Serum creatinine, specific gravity, albumin used. 
Cross-sectional dataset. \\

\cite{15} & 1 & 1 & 0 & 3 &
eGFR and creatinine as top predictors. 
Pre-index features. \\

\cite{2.6} & 1 & 2 & 0 & 4 &
Creatinine top predictor. 
Baseline features prior to outcome. \\

\cite{4.4} & 1 & 0 & 0 & 2 &
ACR used as predictor. 
Cross-sectional NHANES data. \\

\cite{5.4} & 1 & 2 & 0 & 4 &
SCr and SCysC used. 
Baseline recruitment data. \\

\cite{5.6} & 0 & 0 & 0 & 0 &
No diagnostic biomarkers used. 
Predictors before CKD diagnosis. \\

\cite{5} & 1 & 2 & 0 & 4 &
eGFR and UACR predictors. 
Baseline data. \\

\cite{6.3} & 1 & 2 & 0 & 4 &
GFR and creatinine used. 
Cross-sectional dataset. \\

\cite{6.5} & 0 & 0 & 0 & 0 &
No diagnostic predictors. 
Cross-sectional NHANES. \\

\cite{6.6} & 1 & 2 & 0 & 4 &
Serum creatinine used. 
Cross-sectional dataset. \\

\cite{7.4} & 1 & 1 & 0 & 3 &
Blood urea and hemoglobin predictors. 
Cross-sectional dataset. \\

\cite{1.5} & 1 & 2 & 0 & 4 &
Direct: Uses serum creatinine, albumin, specific gravity as selected features, hemoglobin--PCV correlation $\sim 0.90$. 
Cross-sectional UCI data. \\

\cite{1.6} & 1 & 2 & 0 & 4 &
Direct: Serum creatinine listed as primary feature,
Baseline medical test results. \\

\cite{12} & 1 & 2 & 0 & 4 &
Uses serum creatinine, blood pressure, specific gravity,
hemoglobin and PCV in selected features,
cross-sectional clinic data. \\

\cite{13.6} & 1 & 2 & 0 & 4 &
Serum creatinine (sc) and albumin (al) in feature set,
hemoglobin and PCV as top continuous features. \\

\cite{3.5} & 1 & 2 & 0 & 4 &
Explicitly engineers eGFR as novel predictor,
hemoglobin and PCV as significant predictors,
retrospective UCI data. \\

\cite{8.1} & 1 & 2 & 0 & 4 &
Uses serum creatinine and albumin,
hemoglobin as highly important feature, 
traditional diagnostic medical records. \\

\cite{identification} & 0 & 0 & 0 & 0 &
Direct: Excludes creatinine/eGFR/ACR; uses blood routine + urinalysis only. No $|r| \geq 0.90$ correlations identified. 
Retrospective but predictors precede diagnosis. \\

\midrule

\multicolumn{6}{p{\textwidth}}{
\footnotesize \textit{
\cite{1.4} uses S2 model $\rightarrow$ All features except pathology features. 
No temporal leakage was identified across the reviewed studies due to the predominance of cross-sectional datasets lacking explicit temporal ordering.
}} \\

\bottomrule
\end{tabularx}
\end{table}
\noindent
\subsection{\textbf{ Effect of Data Leakage on Model Performance}}

The analysis reveals a systematic relationship between information leakage and excessive model performance in  early CKD prediction studies. By applying the proposed three-part taxonomy which are Direct Leakage (DL), Proxy Leakage (PL) and Temporal Leakage (TL), a consistent pattern emerges across the reviewed literature. Studies that incorporate features directly embedded in diagnostic criteria, such as eGFR, serum creatinine or albuminuria, almost always achieve disproportionately high reported accuracies. This is not surprising, as these variables are not merely predictive signals but are, in fact, definitional components of the outcome itself. Consequently, models in these settings are not learning latent disease patterns but are effectively reconstructing the diagnostic rule. This phenomenon is reflected across a large proportion of studies \cite{1.6,13.6,8.1,6.3,6.6,11.4,5} which receive the maximum leakage score due to the simultaneous presence of direct and proxy leakage. Importantly, even in the absence of temporal leakage, the coexistence of direct and proxy leakage is sufficient to produce near-ceiling performance metrics.

Table \ref{tab:leakage_experiment} demonstrates a continuously increasing trend in reported accuracy as the severity of data leakage increases across models. In the clean configuration (no leakage), models achieved accuracy ranging from 92.50\% (AdaBoost) to 97.50\% (LightGBM). Under the proxy leakage condition, where features correlated with diagnostic criteria, all models exhibited measurable performance gains, with AdaBoost showing the most dramatic improvement of +5.00\% in accuracy and +6.29\% in F1-score. Under the direct leakage condition, where definitional components of CKD were included as predictors, all models achieved near-perfect or perfect accuracy (98.75–100.00\%) and AUC (100.00\%), regardless of their baseline performance. Here, Figure \ref{fig:leakage_pipeline} summarizes the overall workflow of the proposed leakage scoring framework.
\vspace{0.5em}
\noindent

\begin{figure}[htbp]
        \centering
        \includegraphics[
        width=0.78\linewidth,
        keepaspectratio
        ]{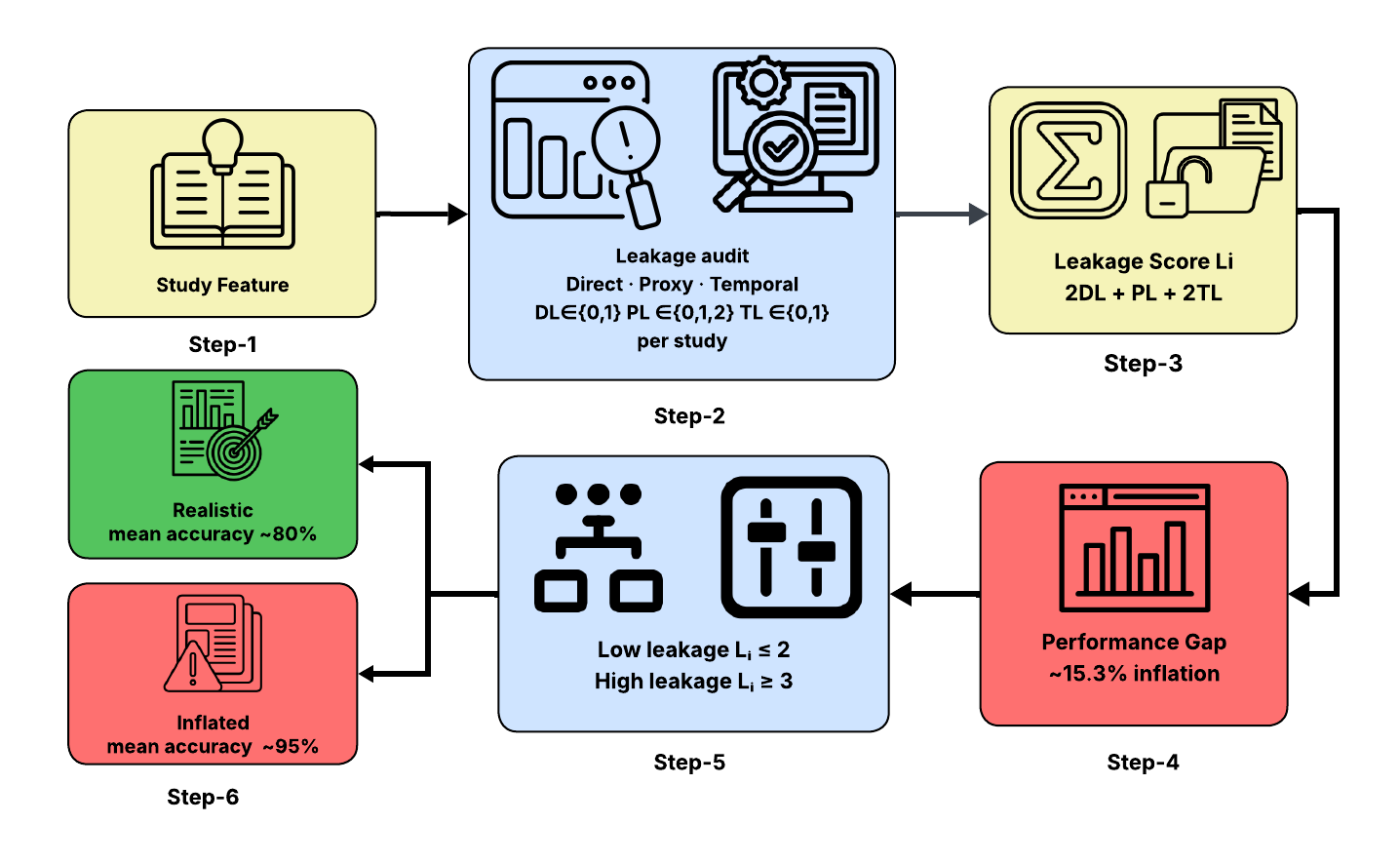}
        \caption{Leakage Scoring Pipeline}
        \label{fig:leakage_pipeline}
\end{figure}

{\subsection{\textbf {Proxy Correlation Undermining Predictive Learning}}}

A detailed examination of proxy leakage shows how machine learning models can generate overly optimistic performance in chronic kidney disease (CKD) prediction tasks. Proxy leakage occurs when non-diagnostic variables are highly correlated ($|r| \geq 0.85$) with established diagnostic biomarkers or clinically redundant with diagnostic biomarkers which enables the model to implicitly reconstruct the target label without learning independent predictive patterns. In clinical datasets, variables such as Blood Urea Nitrogen (BUN), Cystatin C and hemoglobin often demonstrate significant correlations with primary indicators like estimated glomerular filtration rate (eGFR) and serum creatinine. Although these variables represent different clinical measurements, their statistical connections allow models to avoid authentic predictive learning by leveraging overlapping data. This phenomenon is demonstrated in studies such as \cite{7.4} and \cite{15}, where moderate leakage correlates with high accuracy, suggesting that even limited proxy leakage can greatly inflate reported performance.

Importantly, proxy leakage in clinical datasets is not always reflected through high pairwise correlation alone. In several cases (UCI-based studies), individual correlations with diagnostic biomarkers remain moderate ($|r| \approx 0.5$--$0.6$), yet multiple physiologically linked variables (blood urea nitrogen, hemoglobin, packed cell volume) are simultaneously included. These variables collectively encode the same underlying renal dysfunction, forming a multivariate proxy structure. As a result, PL=2 is assigned in such cases to reflect cumulative redundancy, even in the absence of a single dominant correlation.

\begin{figure}[!t]
        \centering
        \includegraphics[
        width=0.78\linewidth,
        keepaspectratio
        ]{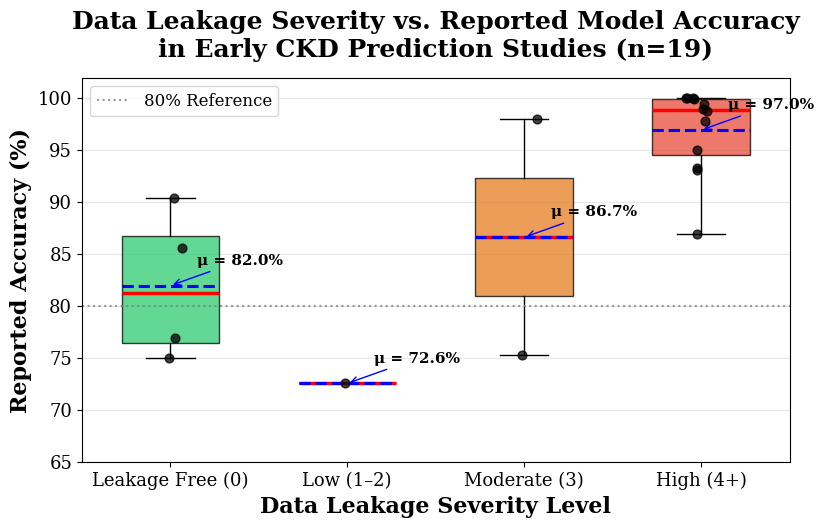}
        \caption{Data Leakage Severity on Reported Accuracy}
        \label{fig:leak level}
\end{figure}
Due to limited public availability of all datasets used in the reviewed studies (Table \ref{tab:leakage_assessment}), correlation analysis was conducted on representative datasets (UCI CKD, Enam Medical, NHANES and Abu Dhabi cohorts). For studies utilizing identical or derived datasets, proxy leakage labels were assigned based on these empirical correlations combined with reported feature sets.

Apart from performance inflation, proxy leakage affects the learning behavior of machine learning models. Tree-based ensemble methods, including XGBoost and Random Forest, are particularly sensitive due to their split-based optimization process. During training, these models select features that maximize information gain and highly correlated predictors are often interchangeable in this process. As a result, proxy variables can repeatedly appear in decision splits, reinforcing the same underlying signal across the ensemble \cite{2.6,5.4,6.6,15,7.4}. This leads to models that appear to learn clinically meaningful relationships but are actually exploiting statistical redundancy. Consequently, variables such as hemoglobin and blood urea nitrogen may receive high importance even though they primarily reflect renal impairment already encoded by creatinine or eGFR \cite{7.4,12,3.5}.

Rankings of feature importance under proxy leakage might not represent independent risk factors but rather serve as substitute indicators of diagnostic criteria. Numerous studies recognize hemoglobin and packed cell volume as important predictors \cite{3.5,12,7.4,8.1}, but these factors are physiologically related to kidney function and highly correlated with creatinine-based metrics, resulting in erroneous conclusions where models seem to indicate causal relationships but actually reflect disease status. This likewise diminishes external validity, since models developed on datasets with robust internal correlations frequently perform poorly on external cohorts having different correlation patterns \cite{3.5,1.5,identification,8.1}. This hybrid formulation ensures that proxy leakage is not underestimated in heterogeneous clinical datasets where statistical correlation alone may fail to capture underlying physiological dependencies.
\begin{table}[t]
\centering
\caption{Performance Comparison}
\label{tab:performance_comparison}

\begin{threeparttable}
\small   

\setlength{\tabcolsep}{25pt}  

\begin{tabular}{lccc}
\toprule
\textbf{Group} & 
\textbf{Papers} & 
\textbf{Mean Acc. (\%)} &  
\textbf{Std. Dev.} \\
\midrule

Low Leakage & 
5 & 
80.16 &  
7.55 \\

High Leakage & 
14 & 
95.48 &  
6.92 \\

\bottomrule
\end{tabular}

\begin{tablenotes}
\footnotesize
\item \textbf{Welch’s t-test:} Mean Diff. = 15.32, SE = 3.85, t = 3.98, $df \approx 7$, p = 0.005, 95\% CI = [6.2, 24.4].
Low leakage if score $\leq 2$ and High leakage if score $\geq 3$.
\item \textbf{Observation:} Statistically significant difference (p = 0.005), suggesting leakage inflates accuracy by (6-24)\%.
\item \textbf{Note:} It should be noted that the relatively small sample size of the leakage-low group (n = 5) limits statistical power. Therefore, while the observed difference is statistically significant (p = 0.005), the result should be validated in larger study cohorts.
\end{tablenotes}

\end{threeparttable}
\end{table}
\vspace{0.5em}
\\
\subsection{\textbf{ Statistical Evidence of Leakage-Induced Inflation}}

The aggregate performance comparison in Table \ref{tab:performance_comparison} provides clear evidence of leakage-induced inflation. This highlights the importance of rigorous leakage auditing before evaluating model performance or reporting results in early CKD prediction. 

The statistical significance (p = 0.005) of Table \ref{tab:performance_comparison} indicates that the observed difference is very unlikely to be explained solely by sampling variability. The consistent inflation of model performance due to leakage indicates a methodological bias instead of authentic predictive capability.

Figure \ref{fig:leak level} represents a gradual rise in reported accuracy as the severity of data leakage increases across multiple studies. Models classified under the high leakage category consistently report inflated performance, indicating potential overestimation of true predictive ability. In contrast, leakage-free studies exhibit relatively lower and more varied accuracy values, indicating more authentic model performance.

Besides statistical significance, the size of the observed difference indicates a considerable practical effect. An average inflation exceeding 15.2\% implies that models utilizing predictors prone to leakage could greatly exaggerate their diagnostic ability in practical scenarios. Importantly, most of the studies where leakage was identified fall into the moderate leakage category, indicating that even small amounts of leakage can influence the reported performance of machine learning models, where inconsistent handling of leakage across studies undermines the reliability of cross-study performance comparisons and further complicates fair benchmarking of models.These findings collectively indicate that to have a fair comparison between the published performance benchmarks, we first need to audit and standardize how leakage is handled.
\begin{table*}[t]
\centering
\small
\setlength{\tabcolsep}{4pt} 
\renewcommand{\arraystretch}{0.96} 
\caption{Frequency, Stability and Consistency of Clinical Predictors Across Reviewed CKD Studies (N=19)}
\label{tab:cross-sec}
\begin{tabular}{p{2.58cm}p{3.3cm}p{1cm}p{1.2cm}p{2.15cm}p{5.8cm}}

\textbf{Predictor Category} & \textbf{Normalized Feature} & \textbf{Studies (n/19)} & \textbf{Stability Score} & \textbf{Consistency Level} & \textbf{References} \\
\hline
\hline
\noalign{\vskip 2pt}
Demographics & Patient Age & 16/19 & 0.84 & High & \cite{1.5,3.5,13.6,identification,12,8.1,1.6,1.4,2.6,4.4,5.4,5.6,5,6.3,6.5,15} \\

Clinical & Blood Pressure & 15/19 & 0.79 & High & \cite{1.4,2.6,4.4,5.6,6.3,6.5,6.6,7.4,1.5,3.5,13.6,12,8.1,1.6,15} \\

Glycemic & Glycemic Control (DM/HbA1c/FBG/BGR) & 14/19 & 0.73 & High & \cite{2.6,4.4,5.4,6.3,6.6,11.4,15,1.5,3.5,13.6,12,8.1,1.6,identification} \\

Hematology & Hemoglobin / Anemia & 13/19 & 0.68 & High & \cite{1.4,6.3,6.6,7.4,11.4,15,1.5,1.6,12,13.6,3.5,8.1,identification} \\

Renal & Serum Creatinine & 13/19 & 0.68 & High & \cite{2.6,4.4,5.4,6.3,6.6,11.4,15,1.5,1.6,3.5,8.1,12,13.6} \\

Renal & Albuminuria (UACR/ACR/Albumin) & 12/19 & 0.63 & High & \cite{1.4,4.4,6.6,7.4,11.4,5,1.5,12,13.6,3.5,8.1,identification} \\

Demographics & Gender / Sex & 11/19 & 0.58 & Moderate & \cite{1.4,2.6,5.4,6.5,1.5,3.5,1.6,12,13.6,identification,8.1} \\

Hematology & Hematology (RBC/RBCC/PCV) & 10/19 & 0.52 & Moderate & \cite{1.4,6.6,7.4,1.5,1.6,12,13.6,3.5,8.1,identification} \\

Medical History & Cardiovascular & 8/19 & 0.42 & Moderate & \cite{1.4,2.6,6.5,1.5,1.6,13.6,3.5,8.1} \\

Urinalysis & Specific Gravity & 8/19 & 0.42 & Moderate & \cite{7.4,11.4,1.5,12,13.6,3.5,8.1,1.6} \\

Renal & BUN / Blood Urea & 8/19 & 0.42 & Moderate & \cite{4.4,7.4,1.5,1.6,13.6,3.5,8.1,11.4} \\

Hematology & WBC Count & 7/19 & 0.37 & Moderate & \cite{1.5,1.6,13.6,3.5,8.1,7.4,identification}\\ 

Electrolyte & Potassium/Sodium & 6/19 & 0.32 & Moderate & \cite{1.5,7.4,1.6,13.6,3.5,8.1} \\

Clinical & Appetite & 6/19 & 0.32 & Moderate & \cite{1.5,1.6,12,13.6,3.5,8.1} \\

Renal & eGFR (as input feature) & 5/19 & 0.26 & Low & \cite{1.1,5,6.3,15,3.5} \\

Clinical & Pus Cell (clumps) & 5/19 & 0.26 & Low & \cite{1.5,1.6,13.6,3.5,8.1} \\

Anthropometric & BMI / Obesity & 4/19 & 0.21 & Low & \cite{1.4,4.4,5.4,6.5} \\

Clinical & Pedal Edema & 4/19 & 0.21 & Low & \cite{1.5,13.6,3.5,8.1} \\

Metabolic & Serum Uric Acid & 3/19 & 0.16 & Low & \cite{4.4,6.5,identification} \\

Lifestyle & Sleep Duration & 3/19 & 0.16 & Low & \cite{1.4,4.4,6.5} \\

Renal & Serum Cystatin C & 2/19 & 0.10 & Low & \cite{5.4,5} \\

Dietary & Dietary Micronutrients & 2/19 & 0.10 & Low & \cite{4.4,6.5} \\

Socioeconomic & Marital Status & 2/19 & 0.10 & Low & \cite{1.4,6.5} \\

Inflammatory & C-Reactive Protein (CRP) & 2/19 & 0.10 & Low & \cite{6.5,5.4} \\

Lifestyle & Physical Activity (MET) & 2/19 & 0.10 & Low & \cite{4.4,6.5} \\

Hematology & Neutrophil Percentage & 1/19 & 0.05 & Low  & \cite{5} \\

Comorbidity & Osteoarthritis Duration & 1/19 & 0.05 & Low & \cite{1.1} \\

Lifestyle & Smokeless Tobacco Use & 1/19 & 0.05 & Low & \cite{1.4} \\

\end{tabular}
\vspace{4pt}
\footnotesize{\textit{Note: Consistency levels are defined as follows — High: 12--19 studies, Moderate: 6--11 studies, Low: 0--5 studies and $N_i = \frac{\text{frequency of predictor } i \text{ across all papers}}{19}$)}}
\end{table*}

\section{Cross-Study Feature Stability Analysis}
The growing use of machine learning models to predict Chronic Kidney Disease (CKD) has resulted in a fragmented and often inconsistent understanding of key factors of the disease. Top predictors of CKD differ across studies, making it difficult to detect characteristics that indicate generalizable risk signals. This inconsistency raises an essential question: Which predictors are actually reliable across different methods and geographies? which result from dataset-specific
traits, methodological biases?

\begin{table}[t]
\centering
\small
\caption{Cross-Dataset Stability of Clinical Predictors}
\label{tab:table_8}
\setlength{\tabcolsep}{4pt}
\renewcommand{\arraystretch}{0.95}

\begin{tabular}{p{4.2cm}p{3.2cm}p{3.3cm}p{4.4cm}}
\textbf{Feature} & \textbf{UCI (n=8)} & \textbf{Non-UCI (n=11)} & \textbf{Avg Cross data Stability} \\
\hline
\hline
\noalign{\vskip 2pt}
Patient Age & 6 & 10 & 0.83 \\
Blood Pressure & 7 & 8 & 0.80 \\
Glycemic Control & 8 & 6 & 0.77 \\
Hemoglobin & 8 & 5 & 0.73 \\
Serum Creatinine & 8 & 5 & 0.73 \\
Albuminuria  & 6 & 6 & 0.65 \\
Gender / Sex & 5 & 6 & 0.59 \\
Hematology  & 6 & 4 & 0.56 \\
Specific Gravity & 6 & 2 & 0.47 \\
BUN / Blood Urea & 6 & 2 & 0.47 \\
Cardiovascular & 5 & 3 & 0.45 \\
WBC Count & 5 & 2 & 0.40 \\
Potassium  & 5 & 1 & 0.36 \\
Appetite & 5 & 1 & 0.36 \\
Pus Cell  & 5 & 0 & 0.31 \\
Pedal Edema & 4 & 0 & 0.25 \\
eGFR (input) & 2 & 3 & 0.22 \\
BMI  & 0 & 4 & 0.18 \\
Serum Uric & 0 & 3 & 0.14 \\
Sleep Duration & 0 & 3 & 0.14 \\
Serum Cystatin C & 0 & 2 & 0.09 \\
Dietary Micronutrients & 0 & 2 & 0.09 \\
Marital Status & 0 & 2 & 0.09 \\
C-Reactive Protein & 0 & 2 & 0.09 \\
Physical Activity & 0 & 2 & 0.09 \\
Neutrophil (\%) & 0 & 1 & 0.05 \\
Osteoarthritis & 0 & 1 & 0.05 \\
Smokeless Tobacco & 0 & 1 & 0.05 \\
\end{tabular}
\end{table}

To address this problem, a systematic cross-study feature stability analysis was performed on nineteen CKD prediction studies. Specifically, 28 standardized attributes  (Table \ref{tab:cross-sec}) were classified into groups. This detailed classification allows for a more refined understanding of feature importance and highlights the interdisciplinary nature of CKD risk modeling.

Each predictor was evaluated based on its frequency of occurrence across the 19 studies and a normalized stability score was computed. High-consistency predictors included patient age, blood pressure, glycemic control, hemoglobin/anemia, serum creatinine and albuminuria. These features cover multiple clinical areas, reinforcing their apparent robustness across diverse datasets and modeling techniques.

\begin{figure}[!b]
    \centering    
    \includegraphics[
    width=0.60\linewidth,
    keepaspectratio
    ]{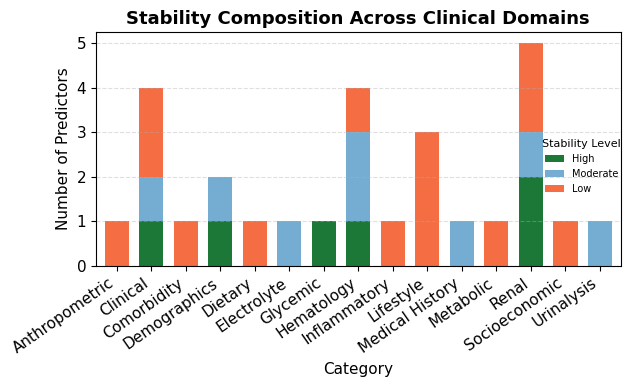}
    \caption{Stacked Bar of Clinical Domains}
    \label{fig:4}
\end{figure}

Figure \ref{fig:4} presents the distribution of predictors across stability levels (high, moderate, low) within each clinical domain. The renal and clinical areas show the highest concentration of predictors overall and contain the majority of high and moderate stability features, reflecting their frequent inclusion in early CKD prediction models. In contrast, domains such as lifestyle, socioeconomic, dietary and inflammatory factors are composed almost entirely of low-stability predictors, indicating restricted usage across various studies. Numerous domains like anthropometric, metabolic and comorbidity factors include just one or two predictors with low stability representation. In general, the figure shows a significant disparity in feature utilization, with CKD prediction studies focusing on a narrow range of clinical and renal factors.

Category-level aggregation further revealed a disproportionate dominance of renal biomarkers in the literature, with the highest cumulative reporting frequency. This overrepresentation continues even with the well- documented risk of data leakage linked with diagnostic variables such as serum creatinine and eGFR. Their frequent inclusion likely reflects their data availability and diagnostic role rather than true predictive value.

The cross-dataset stability analysis of Table \ref{tab:table_8} highlights notable differences between predictors derived from studies using the UCI Chronic Kidney Disease Dataset and those evaluated on independent clinical datasets. Demographic and clinical variables such as patient age and blood pressure demonstrate the highest cross-dataset stability, indicating strong generalizability across diverse study populations and modeling approaches.  
Across multiple studies, Indicators like glycemic and hemoglobin maintain comparatively high stability, indicating their association with CKD prediction \cite{1.4,15}. In general, the results suggest that only a limited set of predictors demonstrates strong cross-population stability, highlighting the necessity for wider dataset variety and standardized feature evaluation for upcoming studies.

\begin{table}[!b]
\centering
\small
\caption{Agreement Statistics for Feature Selection}
\label{tab:freiss_kappa}
\setlength{\tabcolsep}{10pt}
\renewcommand{\arraystretch}{0.9}
\begin{tabular}{lc}
\textbf{Statistic} & \textbf{Value} \\

Number of studies & 19 \\
Number of predictors & 28 \\
Fleiss $\kappa$ & 0.224 \\
Agreement Level & Fair \\

\end{tabular}
\end{table}

Inter-study agreement in feature selection was evaluated using Fleiss' kappa (Table \ref{tab:freiss_kappa}). A $\kappa$ value of 0.224 was obtained, indicating fair agreement among the examined studies. This suggests that although a limited number of predictors (such as age, blood pressure and glycemic indicators) appear consistently across studies, the majority of features are inconsistent. The relatively low k value highlights the disjointed nature of feature selection and supports the observation that many predictors are specific to datasets rather than universally reliable.

In Figure \ref{thresold}, the frequency distribution of all 28 predictors across the 19 reviewed studies is shown, ranked by occurrence. A steep decline in frequency is observed beyond the sixth predictor, with the majority of features appearing in five or fewer studies. This distribution visually reinforces the skewed nature of feature selection in CKD prediction research, where a small number of predictors dominate while most are inconsistent.
\begin{table}[!b]
\centering
\small
\caption{Comparison of feature stability across all studies (n=19) and leakage-aware studies (n=5)}
\setlength{\tabcolsep}{5pt}
\renewcommand{\arraystretch}{1.0}

\begin{tabular}{p{4cm} p{4.40cm} p{5.55cm}}
\textbf{Feature} & \textbf{All Studies Stability (19)} & \textbf{Leakage safe Stability (5)} \\
\hline
\hline
\noalign{\vskip 2pt}
Patient Age & 0.84 & 1.00 \\
Glycemic Control  & 0.73 & 1.00 \\
Blood Pressure & 0.79 & 0.80 \\
Gender / Sex & 0.58 & 0.80 \\
BMI / Obesity & 0.21 & 0.80 \\
Albuminuria & 0.63 & 0.60 \\
Hemoglobin / Anemia & 0.68 & 0.40 \\
Hematology & 0.52 & 0.40 \\
Sleep Duration & 0.16 & 0.40 \\
Tobacco Use & 0.05 & 0.40 \\
Serum Uric Acid & 0.16 & 0.40 \\
Physical Activity & 0.10 & 0.40 \\
Dietary Micronutrients & 0.10 & 0.40 \\
Marital Status & 0.10 & 0.40 \\
Serum Creatinine & 0.68 & 0.20 \\
Cardiovascular History & 0.42 & 0.20 \\
BUN / Blood Urea & 0.42 & 0.20 \\
WBC Count & 0.37 & 0.20 \\
eGFR (as input) & 0.26 & 0.20 \\
C-Reactive Protein & 0.10 & 0.20 \\

\end{tabular}
\label{tab:stability_comparison}
\end{table}

The pareto curve of Figure \ref{fig:dotdotpoints} demonstrates the combined contribution of predictors to total feature usage across different studies. A considerable share of overall usage is represented by high-ranking features as it can be seen from the sharp initial rise in the graph. Specifically, the initial group of predictors accounts for about 80\% of total coverage, illustrating a significant concentration effect. This pattern highlights the dominance of a limited core feature set in CKD prediction models.

\begin{figure}[!t]
    \centering
    \includegraphics[
    width=0.50\linewidth,
    keepaspectratio
    ]{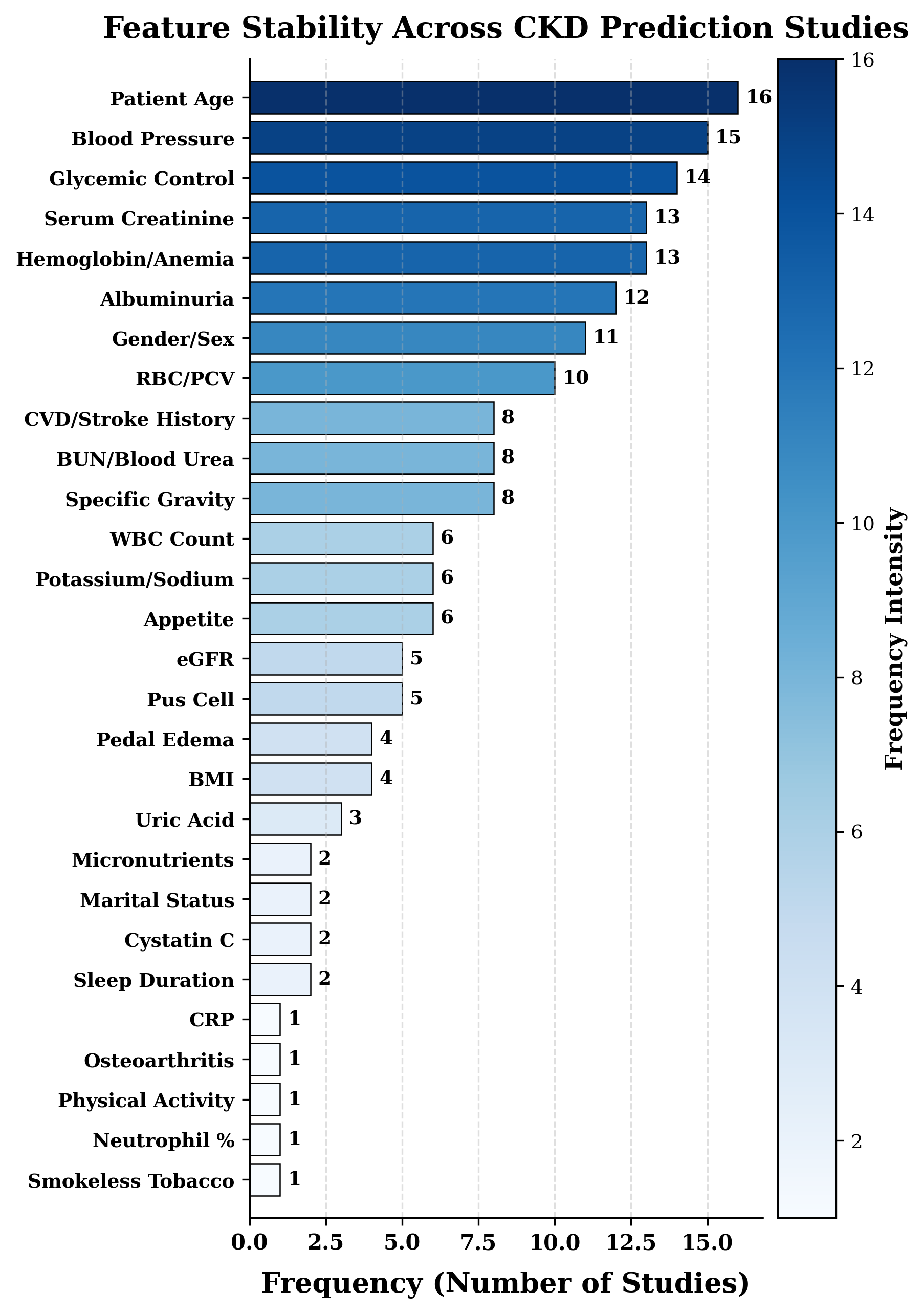}
    \caption{Feature Stability With Frequencies}
    \label{thresold}
\end{figure}
\noindent

\begin{figure}[!t]
    \centering
    \includegraphics[
    width=0.66\linewidth,
    keepaspectratio
    ]{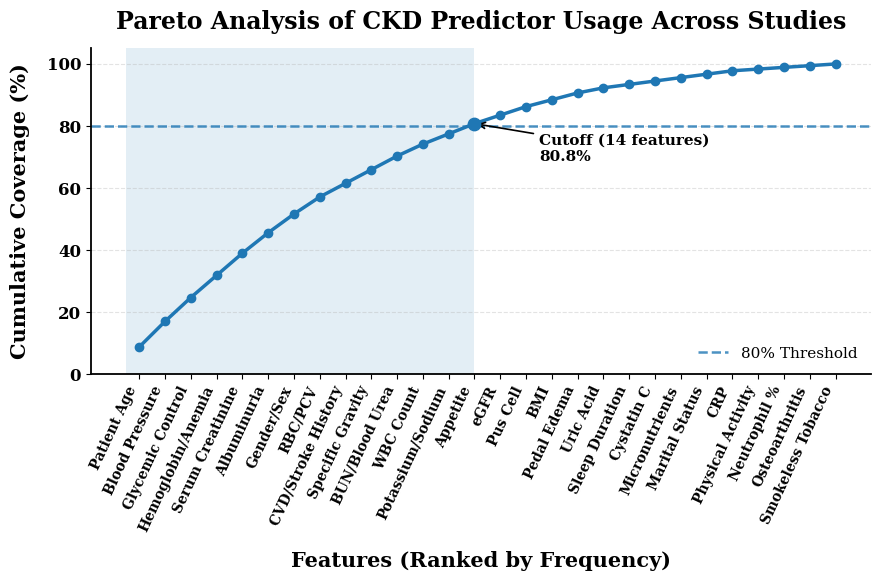}
    \caption{\small Pareto Analysis of CKD Predictor Usage Across Studies}
    \label{fig:dotdotpoints}
\end{figure}

\begin{table}[!t]
\centering
\small
\caption{\centering\\Top 10 Core CKD Predictors After Dataset and Leakage Adjustment}
\label{tab:core-predictors}
\begin{tabular}{p{3.5cm}p{2.8cm}p{2.8cm}p{2.7cm}p{2.6cm}}

\textbf{Predictor} & \textbf{Normal Stability} & \textbf{Cross-Dataset} & \textbf{Leakage-Safe} & \textbf{Average} \\
& \textbf{(n=19)} & \textbf{Stability (n=19)} & \textbf{Stability (n=5)} & \textbf{Stability} \\
\noalign{\vskip 2.5pt}
\hline
\hline
\noalign{\vskip 2pt}
Patient Age & 0.84 & 0.83 & 1.00 & 0.89 \\
Glycemic Control & 0.73 & 0.77 & 1.00 & 0.83 \\
Blood Pressure & 0.79 & 0.80 & 0.80 & 0.80 \\
Gender & 0.58 & 0.59 & 0.80 & 0.66 \\
Albuminuria & 0.63 & 0.65 & 0.60 & 0.63 \\
Hemoglobin & 0.68 & 0.73 & 0.40 & 0.60 \\
Serum Creatinine & 0.68 & 0.73 & 0.20 & 0.54 \\
Hematology (RBC/PCV) & 0.52 & 0.56 & 0.40 & 0.49 \\
BMI & 0.21 & 0.18 & 0.80 & 0.40 \\
BUN & 0.42 & 0.47 & 0.20 & 0.36 \\
\noalign{\vskip2pt}
\multicolumn{5}{p{10cm}}{\footnotesize \textit{ Normal: Consistency across all 19 studies.}} \\
\multicolumn{5}{p{10cm}}{\footnotesize \textit{ Cross-Dataset: UCI vs. Non-UCI generalizability.}} \\
\multicolumn{5}{p{10cm}}{\footnotesize \textit{ Leakage-safe: From Table-\ref{tab:leakage_assessment}.}} \\
\end{tabular}
\end{table}

The analysis of Table \ref{tab:stability_comparison} illustrates a distinct ranking of clinically reliable predictors across the five leakage-safe studies. Patient age and glycemic control emerge as the most stable and universally utilized features, indicating their robust and consistent predictive relevance. Evidence from broad inclusion and higher stability, further supports blood pressure and gender as fundamental predictor. In contrast, variables such as BMI and albuminuria show moderate consistency, suggesting situational relevance across datasets. Several features, including lifestyle factors and  hematological markers, show low stability due to limited inclusion or data availability. Specifically, traditionally robust clinical predictors like serum creatinine and eGFR are often excluded to avoid data leakage.

Table \ref{tab:core-predictors} highlights that patient age, glycemic control and blood pressure are strong indicators, maintaining high performance even after leakage adjustment. Gender, albuminuria and hemoglobin demonstrate moderate to high stability, enhancing their clinical significance. Conversely, the decreasing stability of serum creatinine and hematological parameters under leakage-safe conditions suggests a potential reliance on dataset-specific signals.

Only a limited number of predictors display both conceptual independence from diagnostic standards and methodological strength. Across all three evaluation dimensions, patient age, blood pressure and glycemic control consistently achieve the highest average stability, marking the minimum reliable foundation for CKD risk modeling. Serum creatinine and albuminuria, despite their stability, are directly associated with CKD diagnosis, raising concerns about diagnostic data-leakage. A fragmented research landscape is reflected  in the dominance of low-consistency indicators. Lifestyle, dietary and socioeconomic factors remain largely unexplored, although they offer potential as leakage-free early indicators. Future research should focus on external validation of the stable core predictors and investigate neglected modifiable risk factors. Moreover, Future work should apply standardized methods for feature selection to enhance the robustness and clinical applicability of machine learning models.
\section{Future Directions}
Several clear methodological improvements emerge from this review. Researchers should implement data leakage auditing protocols throughout model development and follow standardized reporting methods aligned with TRIPOD guidelines. Additionally, the field would benefit from benchmark datasets that clearly differentiate diagnostic biomarkers from predictive risk factors, enabling more reliable comparisons among models.
Future work could also benefit from focusing on simple temporal features, such as eGFR decline rate or variability across multiple measurements, to more effectively reflect disease progression without relying on complex sequential models. Furthermore, more reliable detection of data leakage, particularly in resource-limited environments, can be obtained by examining non-renal risk factors like lifestyle choices, socioeconomic status.
\section{Limitations}
This review is subject to several important constraints. One limitation is that correlation analysis covers only four representative datasets, which limits the ability to verify findings from studies that used inaccessible data. Geographical bias presents another concern, as
eight of the nineteen reviewed studies use the CKD dataset from an Indian clinical center. Here, proxy leakage labels for the remaining studies rely on reported feature sets and clinical reasoning. Also, the review excludes relevant recent works as it is limited to studies published up to 2025. Moreover, temporal leakage could not be formally validated because of insufficient timestamp data provided in the examined studies. Additionally, most reviewed studies were evaluated on limited or geographically localized datasets, which may reduce real-world clinical generalizability.
\section{Conclusion}
This systematic review analyzed methods in early CKD prediction using machine learning, with a primary focus on information leakage and cross-study predictor stability. One of the findings shows exaggerated predictive effectiveness due to the inclusion of variables that are closely aligned with CKD diagnostic criteria. This study shows that a considerable amount of reported accuracy likely originates from inherent clinical definitions instead of genuine predictive learning. This finding is demonstrated by applying a structured taxonomy of direct, proxy and temporal leakage. Conversely, studies that avoid leakage-prone features tend to report more moderate but clinically relevant performance, highlighting the necessity of careful feature selection. Furthermore, the cross-study analysis indicates a lack of consistency in the importance of low-leakage predictors, with only a few variables, such as age, blood Pressure and glycemic Control, appearing consistently across multiple datasets. This variability suggests that model behavior often reflects dataset-specific characteristics. Consequently, concerns about generalizability and reproducibility are justified. Although temporal dynamics were not explicitly analyzed, their potential significance in understanding disease progression is recognized as a key area for future research. A Pareto curve analysis further supports the findings as it shows that 80\% of cumulative feature coverage is achieved with 14 variables, highlighting the diminishing marginal utility of adding extra predictors in low-leakage CKD models. Overall, the study highlights the necessity for stricter methodological standards and robust validation frameworks to ensure reliable and clinically significant CKD prediction models.
\noindent
\vspace{0.3em}
\\
\textbf{Author Contributions} M.H. conceived the study, designed the methodology, conducted the systematic review, performed analysis, and interpretation, developed the leakage scoring framework, prepared the figures and tables, and wrote the original manuscript draft. N.K. contributed to the study design, assisted with data interpretation, critically reviewed and revised the manuscript, and provided valuable intellectual input. F.S. contributed to the literature review, assisted with data validation and manuscript revision, and provided constructive feedback throughout the study.  All authors read and approved the final manuscript.
\vspace{0.2em}
\\
\textbf{Corresponding Author} mashrul16hossain@gmail.com (Mashrul Hossain)
\vspace{0.2em}
\\
\textbf{Funding} There was no funding obtained to conduct this research.
\vspace{0.2em}
\\
\textbf{Competing Interest} 
The authors declare there were no competing interests.
\vspace{0.2em}
\\
\textbf{Ethics Statement} This study is a systematic review of previously published studies using publicly available and anonymized datasets. No new human or animal data were collected. Accordingly, no ethical approval or informed consent was required.
\\
\textbf{Data Availability}
No new datasets were generated or analyzed. Used dataset is available on UCI Machine Learning Repository.
\vspace{0.2em}
\\
\textbf{Consent for publication}
All authors have read and approved the final manuscript 
\scriptsize
\bibliographystyle{vancouver}
\bibliography{Reference}
\end{document}